\documentclass[journal]{IEEEtran}
\IEEEoverridecommandlockouts
\usepackage{cite}
\usepackage{amsmath,amssymb,amsfonts,mathtools,bm}
\usepackage{amsthm}
\usepackage{algorithm}
\usepackage{algorithmic}
\usepackage{graphicx}
\usepackage{textcomp}
\usepackage{xcolor,float}
\usepackage{tabularx}

\allowdisplaybreaks
\renewcommand{\baselinestretch}{0.965}

\begin{document}
    
    \title{{ Mobility-Aware Computation Offloading for Swarm Robotics using Deep Reinforcement Learning}}
    
    \author{\IEEEauthorblockN{ Xiucheng Wang $^{\dag}$} and \IEEEauthorblockN{Hongzhi Guo$^{ \dag\dag}$\\}
       $^{\dag}$ School of Telecommunications Engineering, Xidian University, Email: xcwang\_1@stu.xidian.edu.cn.\\
       $^{ \dag\dag}$ Engineering Department, Norfolk State University, Norfolk, VA, 23504, Email: hguo@nsu.edu. 
    }
    
    \maketitle

\IEEEdisplaynontitleabstractindextext

\IEEEpeerreviewmaketitle

\begin{abstract}
    Swarm robotics is envisioned to automate a large number of dirty, dangerous, and dull tasks. Robots have limited energy, computation capability, and communication resources. Therefore, current swarm robotics have a small number of robots which can only provide limited spatio-temporal information. In this paper, we propose to leverage the mobile edge computing to alleviate the computation burden. We develop an effective solution based on a mobility-aware Deep Reinforcement Learning(DRL) model at the edge server side for computing scheduling and resource. Our results show that the proposed approach can meet delay requirements and guarantee computation precision by using minimum robot energy.
\end{abstract}

\begin{IEEEkeywords}
Deep reinforcement learning, edge computing, mobility-aware, swarm robotics, wireless communication.

\end{IEEEkeywords}

\section{Introduction}

A swarm of robots cooperating with each other can automate a large number of dirty, dangerous, and dull tasks such as underwater mining and disaster rescue \cite{yinka2012autonomous}. However, battery-powered robots have limited computation capability, 
and communication resources which significantly reduce their operating range and mission duration. 
To employ more robots in a swarm to obtain richer information, we need to develop energy-efficient swarm robotics 
technologies without compromising computation and communication performance. 

To alleviate robots' computation burden, robots can offload tasks to cloud to get better performance\cite{chinchali2019network}. However, robots cannot always have access to the cloud when they are 
operating in underground mines and underwater. Also, robots demand real-time computation 
to allow them to respond efficiently. The large delay caused by cloud 
computing cannot always meet this requirement. Recently, there are large numbers of research works on mobile edge computing, especially 
the task offloading \cite{8692421}. 
However, most of the existing works focus on applications where only one device is considered and most of them process the task period. Swarm robotics have multiple users with multiple tasks to be offloaded. Also, the mobile edge server may have very limited resources since robots are in extreme environments without well-deployed 
infrastructure. 
Currently, there is no existing solution that jointly consider these challenges in the literature while all robots are moving.

In this paper, we consider swarm robotics to have access to a mobile edge server and they can offload computation 
tasks to the edge server. Robots move 
cooperatively to accomplish tasks in extreme environments while 
satisfying wireless networking requirements such as connectivity and network throughput. In this paper we consider that robots do not 
have high computation capability and cannot make optimal offloading decisions locally, whereas the edge server can make efficient real-time 
offloading scheduling decisions. Generally, the contributions of this paper include: (1) we introduce the mobile edge computing to 
swarm robotics to reduce robot computation energy consumption and computation delay; (2) we develop 
the tasks offloading and scheduling model and prove this problem is NP-hard; (3) we consider robots mobility for offloading and scheduling problems; and (4) we solve the problem by using a Deep Reinforcement Learning (DRL)
model, and we show that the proposed approach can efficiently solve the problem.

\section{System Model and Problem Formulation}
\label{sec:compu_model}
\begin{figure}
  \centering
  \includegraphics[width=0.6\columnwidth]{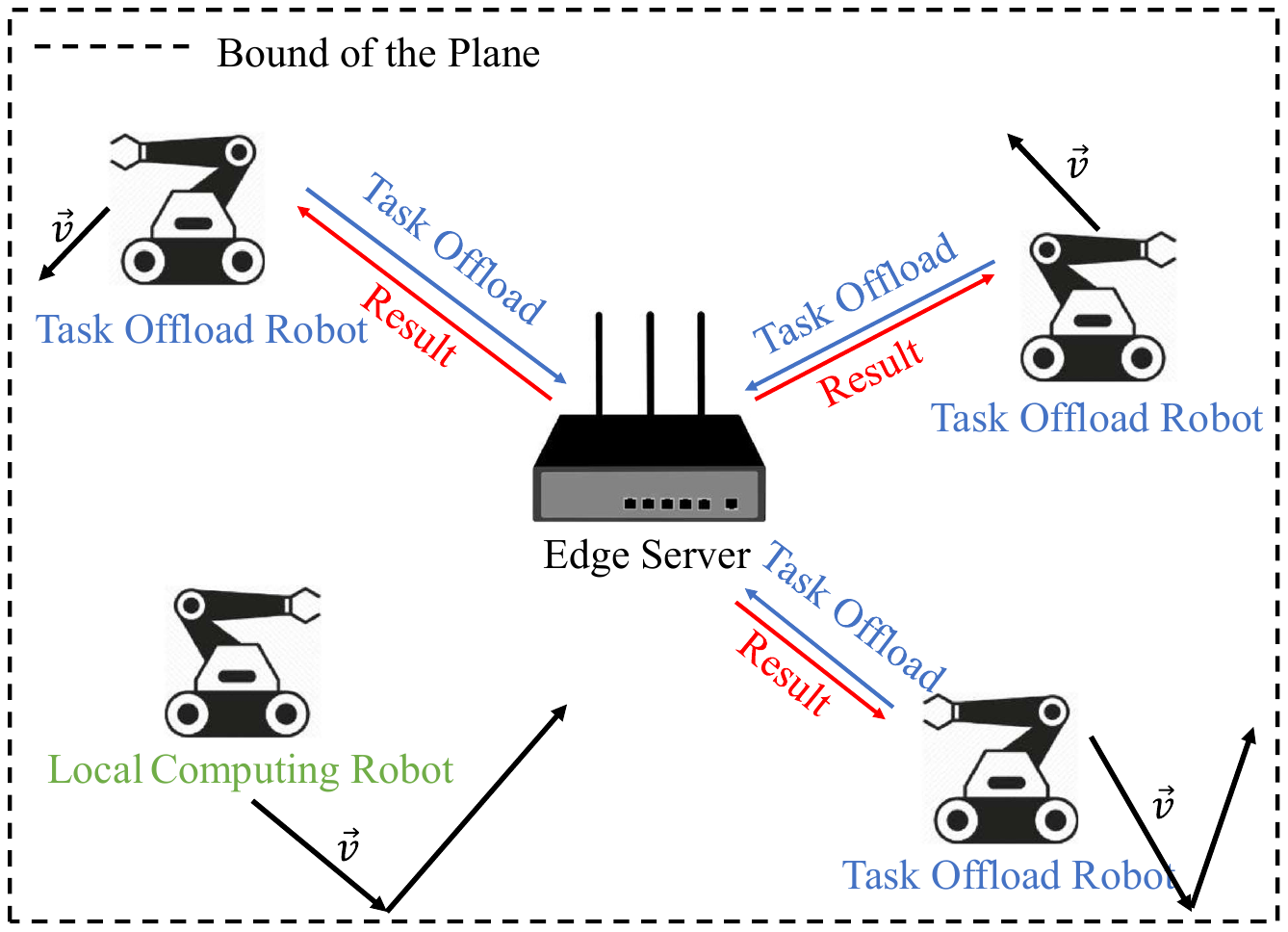}
   \vspace{-5pt}
  \caption {Illustration of the proposed mobile edge computing for swarm robotics.}
  \label{fig:fig1}
   \vspace{-15pt}
\end{figure}

In this section, we first introduce the proposed computation, communication, and energy consumption model for robots and the mobile 
edge server. 
Then, we develop a model for the offloading and service order scheduling problem. 
An illustration of the proposed system is given in Fig.~\ref{fig:fig1}, 
where robots can send task offloading requests to a mobile edge server, and the server can make decisions for each request.

\subsection{System Model}
We consider the CPU frequency of a server as $f_{edg}$ CPU cycles per second, and that of robots as $f_{loc}$.  
There are $N_{rb}$ robots distributed on a finite plane, and the server is stationary in the center of the plane. 
The robots' task is surveying an area while maintaining wireless networking requirements and making optimal swarm motions.

A robot will randomly generate some tasks to process during the movement. We consider that the probability that the robot will 
generate $k$ tasks in any interval $t$, is subject to Poisson random process with intensity $\lambda$, which is
\begin{align}
  P(k) = \frac{(\lambda t)^k}{k!}exp(-\lambda t) ,
\end{align}

Assume that the size of a task is $n$, and $c$ is CPU cycles needed to compute one bit of a task. Then, according to \cite{Energy-Optimal-MCC}, the total consumed energy to compute this task 
locally is $E_{loc}=\gamma n c f_{loc}^{2}$, where $\gamma$ is the effective capacitance coefficient. The time to compute this task is $T_{loc} = {n c}/{f_{loc}}$.
In practice, robots have different types of tasks , e.g., when the robot performs tasks such as fire rescue, it needs 
to respond quickly, while the robot's energy information can be reported periodically in a much slower manner. 
Here, we assume the task is expected to be finished no longer than $d$.

If the robot generates a task within a slot, it will send state information $I=\{d,v,l,n\}$ to the server immediately, $v$ is the movement direction, $l$ is the location of the user. After 
receiving this information, the server decides either compute the task locally or in the server.

Since the application is in extreme environments, the mobile server has limited resources, which is different from that in terrestrial 
environments. Here, we consider the server's processing latency cannot be ignored. The energy consumption of the server 
is neglected.
Assuming the server can only handle one task at a time, and other tasks are waiting in a queue that is used to offer services in order.
When the server agrees to provide services for a task, the robot offloads the task information to the server, and waiting to get the result return. 
Assuming there are up to $M$ tasks waiting to be processed in the waiting queue. 
When there are more than $M$ tasks, the server immediately rejects all service requests. The server must process tasks in waiting queue in sequence.

Initially, all robots are randomly distributed in the plane and move in random directions at a 
constant speed $v$. Robot's mobility is modeled as the robot moves in a straight line at a constant speed according to the initial 
direction of movement, and changes direction when it meets the boundary of the plane. 
According to \cite{UAV} the upload transmission delay for each bit of a task is:
\begin{align}
  T_{e} = \frac{1}{Blog_2(1+P_{tra}h/\sigma^2)}, \label{T_tra}
\end{align}
where $B$ is the bandwidth, $P_{tra}$ is the transmission power, $\sigma^2$ is the power of noise, and the channel gain $h$ follows the free space path loss model. Generally, the size of the computing result of a task is much smaller than the task size and, thus, the latency of the server 
transmitting the computing results to a robot can be ignored. 

According to Equation \eqref{T_tra}, the transmission speed will change as the distance between a robot and a server changes. 
Therefore, due to the movement of robots, the transmission delay $T_{tra}$ is the solution of the integral equation: $\int_{0}^{T_{tra}} Blog_2(1+P_{tra}h/\sigma^2) dt = n$.

\subsection{Problem Formulation}
Assuming all robots generate $N_{t}$ on average tasks per unit time. We define $\bm{X}=\{x_1,x_2,\cdots, x_{N_{t}}\}$ 
as the offloading decision set where $x_i = 0$ means task $i$ is computed locally, $x_i=1$ means the task $i$ is offloaded to the server. 
Also, we use $\bm{J}=\{j_1,j_2\cdots j_{N_{t}}\}$ to denote the queue order of the $N_{t}$ tasks that are processed in the 
server, where $j_i = 0$ means the task will be computed locally, otherwise $j_i$ is the number of computing order in the 
server. Therefore, the energy consumption and delay to complete the $i$th task is:
\begin{align}
  &T_i = x_i(\max\{T_{rea,j_i},T_{tra,i}\} + T_{com,i}) \notag \\
  &\qquad  + (1-x_i) T_{loc,i},\\
  &E_i = x_i(T_{tra,i}P_{tra,i}+(\max\{T_{rea,j_i},T_{tra,i}\} - T_{tra,i})P_{idl,i}) \notag \\
  & \qquad  + (1-x_i) E_{loc,i} , 
\end{align}
where $T_{rea,j_i}$ is the time taken by the server to complete the $j_i$th task, $T_{com,i} = \frac{n_ic}{f_{edg}}$ is the server 
computing time, $T_{tra}$ is the time used to transmit data to the server.
The two goals of the offloading algorithm are to minimize 1) weighted task processing delay, and 2)weighted energy 
consumption to complete the task.
Because the optimization goals are coupled with $\{\bm{X,J}\}$, it is impossible to optimize them independently. We make a 
trade-off among these two goals and define the problem as:
\begin{subequations}
  \begin{align}
    P1: &\underset{{\bm X},{\bm J}}{\text{min}}~ \varpi(\bm{X,J}) =\underset{{\bm X},{\bm J}}{\text{min}}~ \alpha e(\bm{X,J})+\beta t(\bm{X,J}), \label{objective}\\
        &\text{s.t}\;\; j_i\in \{0,1,2\cdots M\},\label{constraint b} \\
        &\quad\;\;\; j_i = 0 \vee  j_i\not= j_k,  \forall j_i\in \bm{J},i\not= k ,\label{constraint c}\\
        &\quad\;\;\; x_i \in \{0,1\}, \label{constraint d}
  \end{align}
\end{subequations}
where $\alpha$ and $\beta$ are weighting parameters, $t(\bm{X,J}) = \sum_{i=1}^{N_{t}} \frac{T_i-d_i}{d_i}$, $e(\bm{X,J}) = \sum_{i=1}^{N_{t}} E_i$. 

The objective function is the weighted sum of the system execution delay and the energy consumption. The constraint \eqref{constraint b} 
guarantees the number of waiting tasks will not exceed $M$, \eqref{constraint c} shows each task's position in the server wait queue is unique, and \eqref{constraint d} guarantees a task can only be processed locally or in the server. However, this problem cannot be solved efficiently since P1 is an NP-hard problem.

To prove it, we consider a special case where $\alpha$ is 0, $\beta$ is 1, and all tasks are generated in the same slot and are all 
offloaded to the server. 
If we consider the transmission delay of a task as the release time of it, $P1$ is equivalent to minimizing $t(\bm{X,J})$ with constraints \eqref{constraint b} and \eqref{constraint c}.
This is a single machine sorting problem with random processing and release time.
Lee\cite{herrbach1990preemptive} proved that it is an NP-hard problem when there are tasks with 
different processing time and release time that need to be processed on even one machine. Therefore, this special case is an NP-hard problem. Since P1 is an NP-hard problem, we propose a solution by using DRL to solve it effectively.

\section{Mobility-Aware Task Scheduling Model}
In this section, we first introduce our proposed scheduling method, and then propose a DRL-based mobility-aware task scheduling model.
\subsection{Task Scheduling Model}
Since the length of the server's waiting queue is $M$, we have to sort up to $M$ tasks. There are total $M!$ choices to sort this problem, which is impossible to be solved by exhaustion search. We consider this problem as a special queuing problem that allows queue cuts.  
If the waiting queue is not full and there are $L$ tasks in the queue ($L<M$), then there are $L+2$ options for the server to handle the task request: 1) reject it; 2) place it in front of the existing $L$ tasks, or 3) place it in the last position of the queue.

If the newly arrived task is not placed at the end of the queue, it will increase the processing time of the existing $L$ tasks. Assuming that the processing delay required by the original $L$ tasks is $T_{o,1},\cdots T_{o,L}$, and the processing delay of the original tasks becomes $T_{n,1},\cdots T_{n,L}$ after a new task arrives, then we define the loss $l_s$ 
caused by the new task to the system as: $l_s = \sum_{i=1}^L \frac{T_{n,i}-T_{o,i}}{d_i}$. When a new task arrives, we can use one-dimensional search to find the best position to minimize the Equation (\ref{objective}), and arrange the new task in this position. 

\begin{algorithm}[!h]
	\caption{Task Scheduling Algorithm}
	\begin{algorithmic}[1]
		\STATE Initialize $Loss=0,order=0,l_s=0$
    \STATE $Loss = \alpha E_{loc}+\beta\frac{d-T_{loc}}{d},$
    \FOR{$i=1$ to $L$}
    \STATE Obtain $l_s,$
    \IF{$Loss >\alpha E_{edg} +\beta (\frac{d-T_{edg}}{d}+l_s)$}
    \STATE $Loss = \alpha E_{edg} +\beta (\frac{d-T_{edg}}{d}+l_s)$
    \STATE $order = i$
		\ENDIF
    \ENDFOR
    \STATE Return $order$
	\end{algorithmic}
\end{algorithm}

\subsection{Mobility-Aware DRL Model}

Since the robots' motion can be regarded as a stable Markov process, so we can use the DRL method to deal with the problem of task scheduling when robots are moving.

\begin{algorithm}[!h]
	\caption{Mobility-Aware DRL Algorithm}
  \begin{algorithmic}[1]
    \STATE Initialize $act$ and $cache$ to 0
		\STATE Get the queue information of the current server and the related information of the new task $D$
    \STATE Input $D$ into the trained neural network and get the output $O$
    \FOR{$i=1$ to $|O|$ (the length of $O$)}
    \IF{$O[i]>cache$}
    \STATE $act = i$
		\ENDIF
    \ENDFOR
    \STATE Return $act$
	\end{algorithmic}
\end{algorithm}

Q-Learning is a classical reinforcement learning method based on state-action pairs (Q-table). An action $a$ in each state $D$ corresponds 
to a $Q(D,a)$, which represents its expected cumulative reward. In Q-Learning method, we aim to maximize the long term reward which is 
$Q(D,a) = \sum_{r=0}^{\infty} r_t$. According to \cite{mnih2015} we can use DRL to approximate this equation as $r_t+\varsigma  \mathop {\max }\limits _{{a}^{\prime }} Q(D^{\prime },a^{\prime };\bm \theta)$
, where $\bm \theta$ is the parameters of the neural network and $\varsigma$ is a discount factor. Next, we introduce the action space, the state space, and the reward.

According to algorithm 1, the size of the action space is equal to $M+1$, and the action space can be represented as $Action = \{0,1,2,\cdot\cdot\cdot,M\}$. We define the state space $D = \{D_d,D_v,D_l,D_n,I\}$, where $D_d$, $D_v$, $D_l$, $D_n$ are the vectors of $d$, $v$, $l$, $n$ for each task in the queue, respectively, and $I$ is the state 
information of the task which is currently under processing, and for the first five vectors, the order of the elements is the same as the order of the tasks in the queue. Since we aim to minimize the average cost of all tasks, if we simply put this task in the first position of the 
waiting queue, it will increase all other tasks processing time in waiting queue which increase the average cost of all tasks. Thus, the reward should also consider the loss $l_s$
of cut in, which is $\sum_{i=k}^L \alpha E_{add,i} + \beta T_{add,i}/U_i$, where $T_{add,i}$ is the extra time the $task_i$
should wait because of the cut in. Therefore, the reward of DRL can be represented as: 
\begin{align}
  R =  -\varpi(\bm{X,J}) - \alpha l_s + \varsigma\mathop {\max }\limits _{{a}^{\prime }} Q(D^{\prime },a^{\prime };\bm \theta),
\end{align}

Next, we show the complexity of developed DRL model with respect to the swarm size. First, the time complexity of our DRL method is $\mathcal{O}(N)$. There are total of 4 layers of fully connected neural networks in DRL, and the number of nodes in each layer is $m_1,m_2,m_3,m_4$. Since the one forward calculation of DRL can be regarded as 3 matrix calculation which is constant, and there are $M$ tasks at most in one slot to process, so there are $M$ times calculation is required at most. Therefore, the time complexity is $\mathcal{O}(N)$. Second, the space complexity of our DRL method is $\mathcal{O}(N)$. The number of weights in the neural network that requires space to be saved is equal to 
  $\sum_{i=1}^3 m_im_{i+1}$. Since $m_2,m_3$ are all constants, $m_1$ is proportional to $M$, $m_4$ is equal to $M + 1$, so the amount of 
  space occupied by the weight increases linearly with the increase of the server's waiting queue length.

\section{Simulation Result}\label{Simulation}
In this section, we evaluate the proposed deep reinforcement learning method by simulation. We assuming there are total $10$ robots moving on a finite square plane with the length of edge being $30$ m, and the length of wait queue $M$ is $10$. The transmission power $P_{tra}$ is $0.05$ W. The CPU frequency of the edge server and robot is $2\times 10^9$ and $2\times 10^8$ cycles per second, respectively.  
Adam optimizer is used to optimize the neural network with learning rate of $0.001$. 
The neural network consists of $4$ fully connected layers. Except for the last layer, which uses Tanh as the activation functions, all other layers use the LeakyReLU activation function with $128$ nodes. 
We assume that the task size $n$ and expectation processing latency $d$ are both obey uniform random distribution which is [120,300]~kb and [0.5,2]~s, respectively. The speed of the user is $2~m/s$, while their motion directions are random. 
Poisson intensity $\lambda$ is $10$. In the simulation, we randomly generated 1 million pieces of data to train our network and other 50 thousand pieces of data to test.

\begin{figure}[t]
  \centering
  \includegraphics[width=0.6\columnwidth]{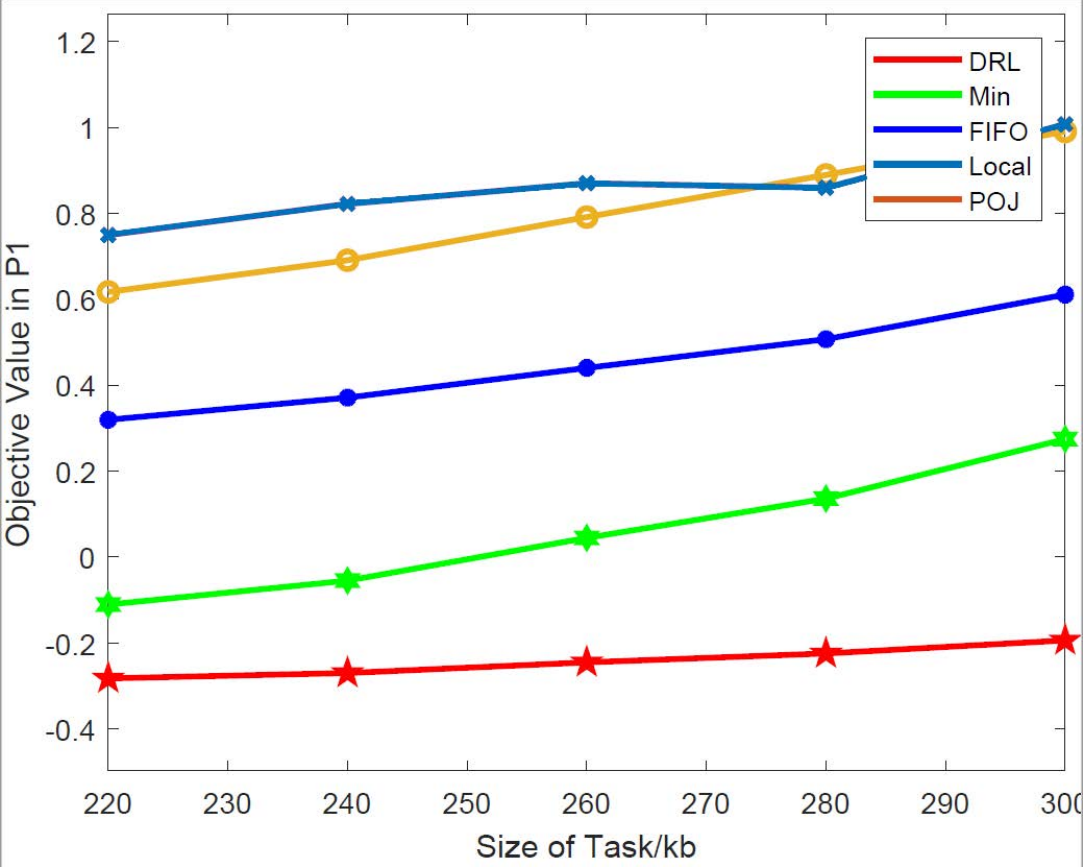}
  \vspace{-5pt}
  \caption {Average performance of the task offloading and scheduling for the given size of computation task.}
  \label{size-o}
   \vspace{-13pt}
\end{figure}

\begin{figure}[t]
  \centering
  \includegraphics[width=0.6\columnwidth]{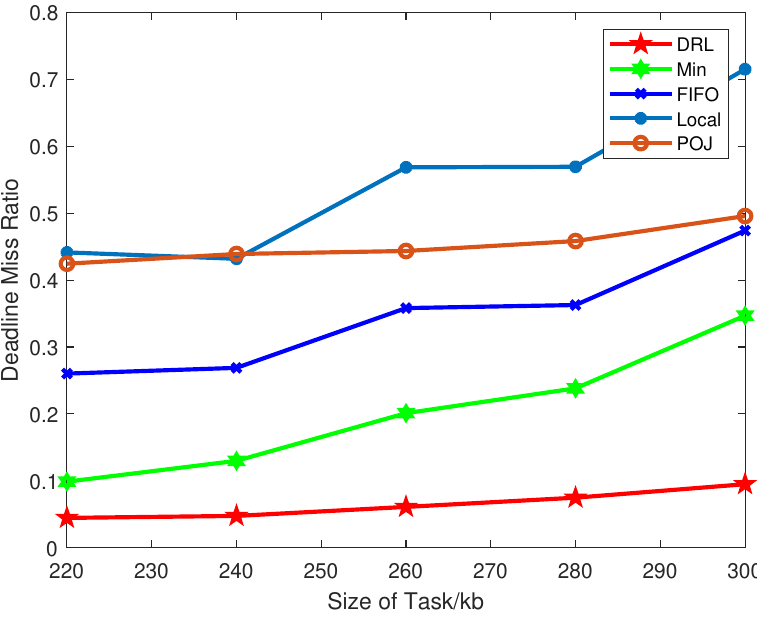}
   \vspace{-5pt}
  \caption {Average performance of the deadline miss ratio for the given size of computation task.}
  \label{size-d}
     \vspace{-10pt}
\end{figure}
\begin{figure}[t]
  \centering
  \includegraphics[width=0.6\columnwidth]{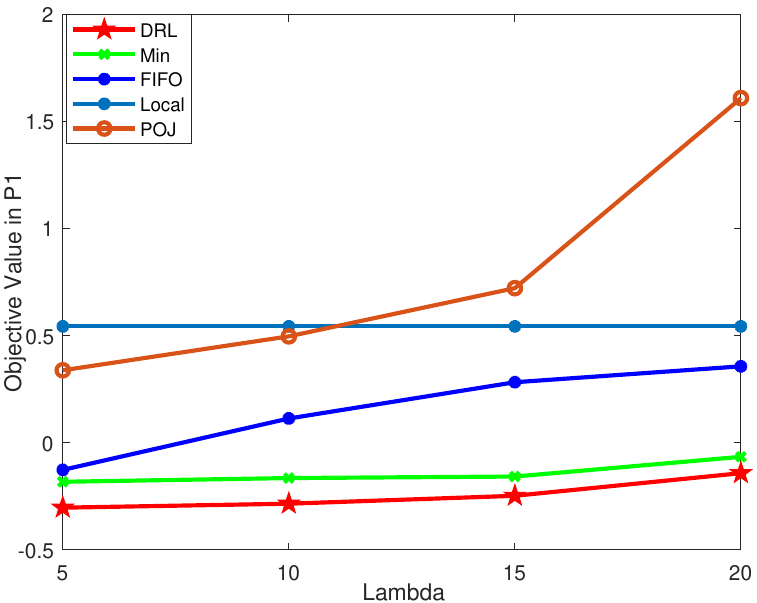}
   \vspace{-5pt}
  \caption {Average performance of the task offloading and scheduling for the average number of tasks generated per second.}
  \label{lambda-o}
     \vspace{-13pt}
\end{figure}

\begin{figure}[t]
  \centering
  \includegraphics[width=0.6\columnwidth]{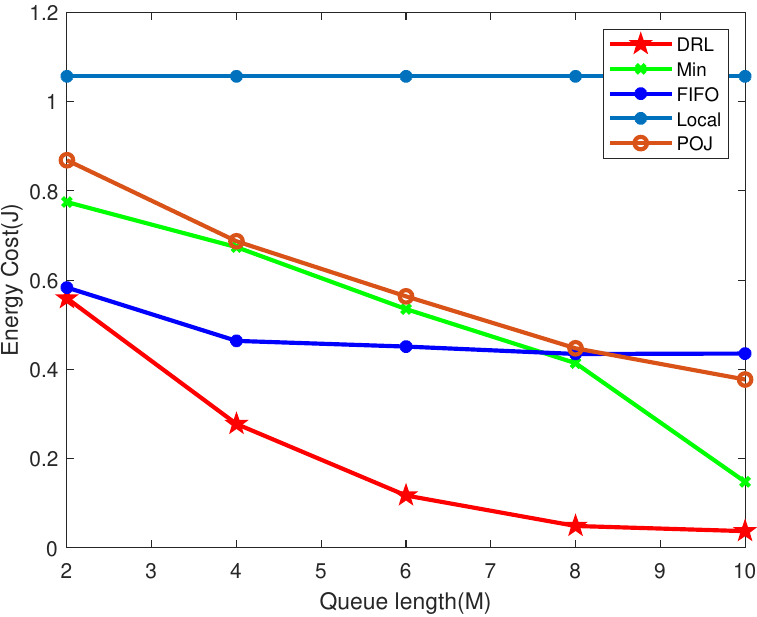}
   \vspace{-5pt}
  \caption {Average performance of the energy cost for the given queue length of edge server.}
  \label{m-e}
     \vspace{-13pt}
\end{figure}

We compare the algorithm proposed in this paper with the offload algorithm proposed by Min in \cite{min} which is also based on 
deep reinforcement learning. We also consider the approaches in \cite{li2018deep} and POJ in \cite{8692421}. Fig.~\ref{size-o} shows the average performance of our proposed DRL method for given task sizes. As the task size increases, the optimization values of all algorithms increases. However, the algorithm proposed in this paper shows excellent stability. Even if the task size increases by 50\%, 
the increase in the optimization target value is very small. Moreover, Fig.~\ref{size-d} shows our method outperforms the POJ\cite{8692421} and 
Min\cite{min} especially in the deadline miss ratio. Since the POJ algorithm needs to periodically wait for the end of a period to process all tasks, when the period value is larger, it will directly cause some tasks to time out when the task starts to be processed. If the period value is small, it will 
lead to a decrease in algorithm performance and an increase in task processing delay. This also shows the importance of real-time processing rather than periodic processing when the task generation process conforms to the Poisson process.

Since the probability of each robot generating task is subject to a Poisson random process with the same parameters, studying the 
influence of the number of robots on performance is equivalent to studying the influence of Poisson strength on performance and, thus, we show the influence of $\lambda$.
Fig.~\ref{lambda-o} shows as the $\lambda$ increasing, the optimization target value increases, and the 
algorithm proposed in this paper and the algorithm in \cite{min} have good stability. The POJ\cite{8692421} 
algorithm increases rapidly after $\lambda$ is greater than 15, which is mainly because the POJ algorithm 
requires periodic processing tasks. When the average number of tasks generated per unit time increases, more 
tasks cannot be processed in time, thereby increasing the average processing delay of tasks. 
This shows that when the rate of task generation in the system is high, the time delay caused by the 
periodic processing of the algorithm itself cannot be ignored, and it also confirms the necessity of 
real-time processing of tasks.

We further analyzed the influence of the edge server waiting queue length on task processing. Fig.~\ref{m-e} shows, as $M$ decreases, 
the energy required for task processing increases. This is because as $M$ decreases, when more tasks are generated, the server queue is 
full, which will cause more tasks to be directly denied service, thereby performing local computing.

\section{Conclusion}
Swarm robotics in extreme environments can automate a large number of important applications. To use cheap tiny robots, their 
computation and energy limitations have to be addressed. This paper proposes an edge computing solution using a mobile server for swarm 
robotics. We developed a deep reinforcement learning decision model and numerically evaluated its performance. The results show that the 
proposed approach can reduce energy consumption while meeting the requirements of computation latency.

\ifCLASSOPTIONcaptionsoff
  \newpage
\fi

\bibliography{ref}
\bibliographystyle{IEEEtran}

\end{document}